\documentclass[runningheads]{llncs}
\usepackage[T1]{fontenc}

\usepackage{graphicx}
\usepackage{subfigure}
\usepackage{amsmath}
\usepackage{amssymb}
\usepackage{paralist}
\usepackage{url}
\usepackage{paralist}
\usepackage{url}
\usepackage{footnote}
\usepackage{algorithmic}
\usepackage{algorithm}
\usepackage{xspace}
\usepackage{graphicx,subfigure}
\usepackage{url}
\usepackage{wrapfig}
\usepackage{caption}

\newcommand{\plr}{\text{plr}\xspace{}}
\newcommand{\hic}{\text{hic}\xspace{}}

\newcommand{\nd}{\text{NormalDistribution}\xspace{}}
\DeclareMathOperator*{\argmmax}{arg\,max}

\usepackage{cite}

\usepackage{xcolor}

\newcommand{\mysubsection}[1]{\medskip\noindent\textbf{#1}}

\begin{document}
\title{RoMA: a Method for Neural Network Robustness	Measurement and Assessment
	}
%
      \author{Natan Levy
        \and
	Guy Katz
      }

\authorrunning{N. Levy and G. Katz}
\titlerunning{RoMA: Robustness Measurement and Assessment}
%

 \institute{The Hebrew University of Jerusalem, Jerusalem, Israel\\
   \email{\{natan.levy1, g.katz\}@mail.huji.ac.il}}

 \toctitle{RoMA: a Method for Neural Network Robustness Measurement
   and Assessment}
 \tocauthor{Natan~Levy and Guy~Katz}

\maketitle	
\begin{abstract}
  Neural network models have become the leading solution for a large
  variety of tasks, such as classification, natural language
  processing, and others. However, their reliability is heavily
  plagued by \emph{adversarial inputs}: inputs generated by adding
  tiny perturbations to correctly-classified inputs, and for which the
  neural network produces erroneous results. In this paper, we
  present a new method called \emph{Robustness Measurement and
  Assessment} (\emph{RoMA}), which measures the robustness of a
  neural network model against such adversarial inputs. Specifically,
  RoMA determines the probability that a random input perturbation
  might cause misclassification. The method allows us to provide
  formal guarantees regarding the expected frequency of errors that a
  trained model will encounter after deployment. The type of
  robustness assessment afforded by RoMA is inspired by
  state-of-the-art certification practices, and could constitute an
  important step toward integrating neural networks in safety-critical
  systems.

\keywords{Neural Networks \and Adversarial Examples \and Robustness \and Certification.}
\end{abstract}

\section{Introduction}
\label{INTRODUCTION}

In the passing decade, deep neural networks (DNNs) have emerged as one
of the most exciting developments in computer science, allowing
computers to outperform humans in various classification
tasks. However, a major issue with DNNs is the existence of
\emph{adversarial inputs}~\cite{GoShSz14}: inputs that are very close
(according to some metrics) to correctly-classified inputs, but which
are misclassified themselves. It has been observed that many
state-of-the-art DNNs are highly vulnerable to adversarial
inputs~\cite{CaWa17}.

As the impact of the AI revolution is becoming evident, regulatory
agencies are starting to address the challenge of integrating DNNs
into various automotive and aerospace systems --- by forming
workgroups to create the needed guidelines. Notable examples in the
European Union include SAE G-34 and EUROCAE WG-114~\cite{PeTh20,ViGaObOb21}; and the European Union Safety Agency (EASA), which is responsible for civil aviation safety, and which has published a road
map for certifying AI-based systems~\cite{EuUnAvSaAg20}. These
efforts, however, must overcome a significant gap: on one hand, the
superior performance of DNNs makes it highly desirable to incorporate
them into various systems, but on the other hand, the DNN's intrinsic
susceptibility to adversarial inputs could render them unsafe. This
dilemma is particularly felt in safety-critical systems, such as
automotive, aerospace and medical devices, where regulators and public
opinion set a high bar for reliability.

In this work, we seek to begin bridging this gap, by devising a
framework that could allow engineers to \emph{bound and mitigate} the
risk introduced by a trained DNN, effectively containing the
phenomenon of adversarial inputs. Our approach is inspired by common
practices of regulatory agencies, which often need to certify various
systems with components that might fail due to an unexpected hazard. A
widely used example is the certification of jet engines, which are
known to occasionally fail. In order to mitigate this risk,
manufacturers compute the engines' \emph{mean time between failures}
(\emph{MTBF}), and then use this value in performing a safety analysis that can eventually justify the safety
of the jet engine system as a whole~\cite{LaNi11}. For example, 
federal agencies guide that the
probability of an extremely improbable failure conditions event per operational hour
should not exceed $10^{-9}$~\cite{LaNi11}. To perform a similar process
for DNN-based systems, we first need a technique for accurately
bounding the likelihood of a failure to occur --- e.g., for measuring
the probability of encountering an adversarial input.

In this paper, we address the aforesaid crucial gap, by introducing a
straightforward and scalable method for measuring the probability that
a DNN classifier misclassifies inputs. The method, which we term
\emph{Robustness Measurement and Assessment} (\emph{RoMA}), is
inspired by modern certification concepts, and operates under the
assumption that a DNN's misclassification is due to some internal
malfunction, caused by random input perturbations (as opposed to
misclassifications triggered by an external cause, such as a malicious
adversary). A random input perturbation can occur naturally as part
of the system's operation, e.g., due to scratches on a camera lens or
communication disruptions. Under this assumption, RoMA can be used to
measure the model's robustness to randomly-produced adversarial
inputs. 

RoMA is a method for estimating rare events in a large population ---
in our case, adversarial inputs within a space of inputs that are
generally classified correctly. When these rare events (adversarial
inputs) are distributed normally within the input space, RoMA performs
the following steps: it
\begin{inparaenum}[(i)]
\item samples a few hundred random input points;
  \item measures the
    ``level of adversariality'' of each such point; and
    \item uses the normal distribution
function to evaluate the probability of encountering an adversarial input
within the input space.
\end{inparaenum}
Unfortunately, adversarial inputs are often \emph{not} distributed
normally. To overcome this difficulty, when RoMA detects this case it
first applies a statistical \emph{power transformation}, called
Box-Cox~\cite{BoCo82}, after which the distribution often becomes
normal and can be analyzed. The Box-Cox transformation is a
widespread method that does not pose any restrictions on the DNN in
question (e.g., Lipschitz continuity, certain kinds of activation
functions, or specific network topology). Further, the method does not
require access to the network's design or weights, and is thus
applicable to large, black-box DNNs.

We implemented our method as a proof-of-concept tool, and evaluated it
on a VGG16 network trained on the CIFAR10 data set.
Using RoMA, we were able to show that, as expected, a higher number of
epochs (a higher level of training) leads to a higher robustness
score. Additionally, we used RoMA to measure how the model's
robustness score changes as the magnitude of allowed input perturbation
is increased. Finally, using
RoMA we found that the \emph{categorial robustness} score of a DNN,
which is the robustness score of inputs labeled as a particular
category, \emph{varies significantly} among the different categories.

To summarize, our main contributions are:
\begin{inparaenum}[(i)]
	\item introducing RoMA, which is a new and scalable method for
		measuring the robustness of a DNN model, and which can be
		applied to black-box DNNs;
	\item using RoMA to measure the effect of additional training
		on the robustness of a DNN model;
	\item using RoMA to measure how a model's robustness changes
		as the magnitude of input perturbation increases; and
	\item formally computing categorial robustness scores,
		and demonstrating that they can differ significantly between labels.
\end{inparaenum}

\mysubsection{Related work.}
The topic of statistically evaluating a model's adversarial robustness
has been studied extensively. State-of-the-art
approaches~\cite{HuHuHuPe21,CoRoZi19} assume that the confidence
scores assigned to perturbed images are normally distributed, and
apply \emph{random sampling} to measure robustness. However, as we
later demonstrate, this assumption often does not hold. Other
approaches~\cite{WeRaTePa18,TiFuRo2021,MaNoOr19} use a sampling method
called \emph{importance sampling}, where a few bad samples with large
weights can drastically throw off the estimator. Further, these
approaches typically assume that the network's output is
Lipschitz-continuous. Although RoMA is similar in spirit to these
approaches, it requires no Lipschitz-continuity, does not
assume a-priori that the adversarial input confidence scores are
distributed normally, and provides rigorous robustness guarantees.

Other noticeable methods for measuring robustness include
formal-verification based approaches~\cite{KaBaDiJuKo17,
  KaBaDiJuKo21}, which are exact but which afford very limited
scalability; and approaches for computing an estimate bound on the
probability that a classifier's margin function exceeds a given
value~\cite{WeChNgSqBoOsDa19,AnSo20,DvGaFaKo18}, which focus on
worst-case behavior, and may consequently be inadequate for regulatory
certification. In contrast, RoMA is a scalable method, which focuses
on the more realistic, average case.

\section{Background}
\label{Background}
\mysubsection{Neural Networks.} A neural network $N$ is a function
$N: \mathbb{R}^n \rightarrow \mathbb{R}^m$, which maps a real-valued
input vector $ \vec{x} \in \mathbb{R}^n$ to a real-value output vector
$\vec{y} \in \mathbb{R}^m$. For classification networks, which is our
subject matter, $\vec{x}$ is classified as label $l$ if $y$'s $l$'th
entry has the highest score; i.e., if $\argmmax(N(\vec{x}))=l$.

\mysubsection{Local Adversarial Robustness.} The local adversarial
robustness of a DNN is a measure of how resilient that network is
against adversarial perturbations to specific inputs. More
formally~\cite{BaIoLaVyNoCr16}:

\begin{definition}
	\label{definition1}
	A DNN $N$ is $\epsilon$-locally-robust at input point $\vec{x_0}$ iff
	\[
	\forall \vec{x}.
	\displaystyle || \vec{x} -\vec{x_0} ||_{\infty} \le \epsilon 
	\Rightarrow \argmmax(N(\vec{x})) = \argmmax(N(\vec{x_0})) 
	\]
\end{definition}

Intuitively, Definition~\ref{definition1} states that for input vector
$\vec{x}$, which is at a distance at most $\epsilon$ from a fixed input
$\vec{x_0}$, the network function assigns to $\vec{x}$ the same label that it assigns
to $\vec{x_0}$ (for simplicity, we use here the $L_\infty$ norm, but other
metrics could also be used). When a network is \emph{not}
$\epsilon$-local-robust at point $\vec{x_0}$, there exists a point $\vec{x}$
that is at a distance of at most $\epsilon$ from $\vec{x_0}$, which is
misclassified; this $\vec{x}$ is called an \emph{adversarial input}. In
this context, \emph{local} refers to the fact that $\vec{x_0}$ is
fixed. 

\mysubsection{Distinct Adversarial Robustness.} Recall that the label
assigned by a classification network is selected according to its
greatest output value. The final layer in such networks is usually a
softmax layer, and its outputs are commonly interpreted as confidence scores assigned to each of the possible labels.\footnote{The term
\emph{confidence} is sometimes used to represent the reliability of
the DNN as a whole; this is not our intention here.} We use
$c(\vec{x})$ to denote the highest confidence score,
i.e. $c(\vec{x})=max(N(\vec{x}))$.

We are interested in an adversarial input $\vec{x}$ only if it is
\emph{distinctly} misclassified~\cite{LaNi11}; i.e., if $\vec{x}$'s
assigned label receives a significantly higher score than that of the
label assigned to $\vec{x_0}$. For example, if
$\argmmax(N(\vec{x_0}))\neq \argmmax(N(\vec{x}))$, but
$c(\vec{x})=20\%$, then $\vec{x}$ is not distinctly an adversarial
input: while it is misclassified, the network assigns it an extremely low
confidence score. Indeed, in a safety-critical setting, the system is expected
to issue a warning to the operator when it has such low confidence in
its classification~\cite{MiKwGa2018}. In contrast, a case where
$c(\vec{x})=80\%$ is much more distinct: here, the network gives an
incorrect answer with high confidence, and no warning to the operator
is expected. We refer to inputs that are
misclassified with confidence greater than some threshold $\delta$ as
\emph{distinctly adversarial inputs}, and refine
Definition~\ref{definition1} to only consider them, as follows:

\begin{definition}
	\label{definition2}
	
	A DNN $N$ is ($\epsilon,\delta$)-distinctly-locally-robust at input point $\vec{x_0}$,
	iff
	\begin{align*}
		&\forall \vec{x}.\ 
		\displaystyle || \vec{x} -\vec{x_0} ||_{\infty}  \le \epsilon  
		\Rightarrow 	\big( \argmmax(N(\vec{x})) = \argmmax(N(\vec{x_0})) \big) \vee (c(\vec{x})<\delta )
	\end{align*}
\end{definition}
Intuitively, if the definition does not hold then there exists a
(distinctly) adversarial input $\vec{x}$ that is at most $\epsilon$ away
from $\vec{x_0}$, and which is assigned a label different than that of
$\vec{x_0}$ with a confidence score that is at least $\delta$.

\section{The Proposed Method}
\label{Method}
\subsection{Probabilistic Robustness}
Definitions~\ref{definition1} and~\ref{definition2} are geared for an
external, malicious adversary: they are concerned with the existence
of an adversarial input. Here, we take a
different path, and follow common certification methodologies that
deal with internal malfunctions of the
system~\cite{FAA93}. Specifically, we focus on ``non-malicious
adversaries'' --- i.e., we assume that perturbations occur naturally,
and are not necessarily malicious. This is represented by assuming
those perturbations are randomly drawn from some distribution. We argue
that the non-malicious adversary setting is more realistic for
widely-deployed systems in, e.g., aerospace, 
which are expected to operate at a large scale and over a prolonged
period of time, and are more likely to encounter randomly-perturbed
inputs than those crafted by a malicious adversary.

Targeting randomly generated adversarial inputs requires extending
Definitions~\ref{definition1} and~\ref{definition2} into a
probabilistic definition, as follows:

\begin{definition}
	\label{definition3}
	
	The
	$(\delta,\epsilon)$-probabilistic-local-robustness
	score of a DNN $N$ at input point $\vec{x_0}$, abbreviated
	$\plr{}_{\delta,\epsilon}(N,\vec{x_0})$, is defined as:
	\begin{align*}
		\plr{}_{\delta,\epsilon}&(N,\vec{x_0})
		\triangleq 
		P_{x:  \lVert \vec{x} -\vec{x_0} \rVert_\infty \le \epsilon}
		[(\argmmax(N(\vec{x})) = 
		\argmmax(N(\vec{x_0}))
		\lor c(\vec{x}) < \delta)]
	\end{align*}
\end{definition}
Intuitively, the definition measures the probability that an input
$\vec{x}$, drawn at random from the $\epsilon$-ball around
$\vec{x_0}$, will either have the same label as $\vec{x_0}$ or, if it
does not, will receive a confidence score lower than $\delta$ for its
(incorrect) label.

A key point is that probabilistic robustness, as defined in
Definition~\ref{definition3}, is a scalar value: the closer this value
is to 1, the less likely it is a random perturbation to $\vec{x_0}$
would produce a distinctly adversarial input. This is in contrast to
Definitions~\ref{definition1} and~\ref{definition2}, which are Boolean
in nature. We also note that the probability value in
Definition~\ref{definition3} can be computed with respect to values of
$\vec{x}$ drawn according to any input distribution of interest. For
simplicity, unless otherwise stated, we assume that $\vec{x}$ is
drawn uniformly at random. 

In practice, we propose to compute
$\plr{}_{\delta,\epsilon}(N,\vec{x})$ by first computing the
probability that a randomly drawn $\vec{x}$ \emph{is} a distinctly adversarial
input, and then taking that probability's complement.
Unfortunately, directly bounding the probability of randomly encountering an
adversarial input, e.g., with the Monte Carlo or Bernoulli methods~\cite{Ha13}, is not feasible due to the typical extreme sparsity
of adversarial inputs, and the large number of samples required to
achieve reasonable accuracy~\cite{WeRaTePa18}. Thus, we require a
different statistical approach to obtain this measure, using only a
reasonable number of samples. We next propose such an approach.

\subsection{Sampling Method and the Normal Distribution}
\label{NormalDistribution}

Our approach is to measure the probability of randomly encountering an
adversarial input, by examining a finite set of perturbed samples
around $\vec{x_0}$. Each perturbation is selected through \emph{simple
random sampling}~\cite{Ta16} (although other sampling methods can be
used), while ensuring that the overall perturbation size does not exceed the given
$\epsilon$. Next, each perturbed input $\vec{x}$ is passed through the
DNN to obtain a vector of confidence scores for the possible output
labels. From this vector, we extract the \emph{highest incorrect
  confidence} (\emph{$\hic{}$}) score:
\[
\hic(\vec{x}) = \max_{i\neq \argmmax(N(\vec{x_0}))} \{N(\vec{x})[i]\}
\]
which is the highest confidence score assigned to an \emph{incorrect}
label, i.e., a label different from the one assigned to $\vec{x_0}$.
Observe that input $\vec{x}$ is distinctly adversarial if and only if
its $\hic{}$ score exceeds the $\delta$ threshold.

The main remaining question is how to extrapolate from the collected
$\hic{}$ values a conclusion regarding the $\hic{}$ values in the
general population. The normal distribution is a useful notion in
this context: if the $\hic{}$ values are distributed normally (as
determined by a statistical test), it is straightforward to obtain
such a conclusion, even if adversarial inputs are scarce.

\begin{figure}[htp]
	\centering
	
	\includegraphics[trim = 2cm 21cm 10cm 3.1cm,clip,width=.75\textwidth]{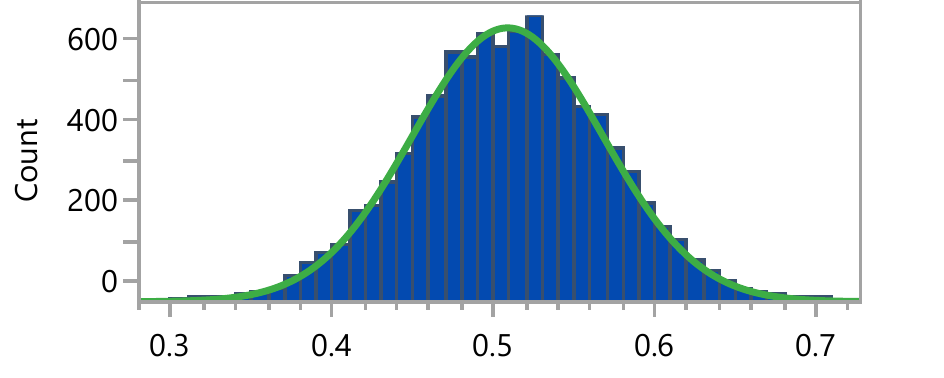} 
	\caption{A histogram depicting the highest incorrect
		confidence ($\hic{}$) scores assigned to each of
		10,000 perturbed inputs. These scores are normally
		distributed.}
	\label{DistributeNarmally}
\end{figure}

To illustrate this process, we trained a VGG16 DNN model (information
about the trained model and the dataset appears in
Section~\ref{Evaluation}), and examined an arbitrary point
$\vec{x_0}$, from its test set. We randomly generated 10,000 perturbed
images around $\vec{x_0}$ with $\epsilon = 0.04$, and ran them through the DNN. For each output vector obtained this way we collected the
$\hic{}$ value, and then plotted these values as the blue histogram in
Figure~\ref{DistributeNarmally}. The green curve represents the normal
distribution. As the figure shows, the data is normally distributed;
this claim is supported by running a ``goodness-of-fit'' test
(explained later).

Our goal is to compute the probability of a fresh, randomly-perturbed
input to be distinctly misclassified, i.e. to be assigned a $\hic{}$ score that
exceeds a given $\delta$, say $60\%$. For data distributed normally, as
in this case, we begin by calculating the \emph{statistical standard
	score} (\emph{Z-Score}), which is the number of standard deviations
by which the value of a raw score exceeds the
mean value. Using the Z-score, we can compute the probability of the event using the Gaussian function. In our case, we get
$ \hic(\vec{x}) \sim \mathcal{N} ( \mu=0.499 , \Sigma=0.059^2)$, where
$\mu$ is the average score and $\Sigma$ is the variance. The 
Z-score is
$\frac{\delta-\mu}{\sigma} = \frac{0.6-0.499}{0.059}=1.741$, where
$\sigma$ is the standard deviation. Recall that our goal is to
compute the $\plr{}$ score, which is the probability
of the $\hic{}$ value not exceeding $\delta$; and so we obtain that:
\begin{align*}
	\plr{}_{0.6,0.04}(N,\vec{x_0}) &= \nd(\text{Z-score}) \\ &= \nd(1.741) \\
	&= \frac{1}{\sqrt{2\pi}}\int_{-\infty}^{t=1.741}e^\frac{-t^2}{2} dt = 0.9591
\end{align*}
We thus arrive at a probabilistic local robustness score of
$95.91\%$.

Because our data is obtained empirically, before we can apply the
aforementioned approach we need a reliable way to determine whether
the data is distributed normally. A \emph{goodness-of-fit} test is a
procedure for determining whether a set of $n$ samples can be
considered as drawn from a specified distribution. A common
goodness-of-fit test for the normal distribution is the
Anderson-Darling test~\cite{An11}, which focuses on samples in the
tails of the distribution~\cite{BeKoSc21}. In our evaluation, a
distribution was considered normal only if the Anderson-Darling test
returned a score value greater than $\alpha=0.15$, which is considered
a high level of significance --- guaranteeing that no major deviation
from normality was found.

\subsection {The Box-Cox Transformation}
\label{Box-Cox}

\begin{figure}[h!]
  \centering
		\subfigure[]
		{\label{DistributionBefore}
		\includegraphics[trim = 2cm 21cm 10cm
		3.1cm,clip,width=.7\textwidth]{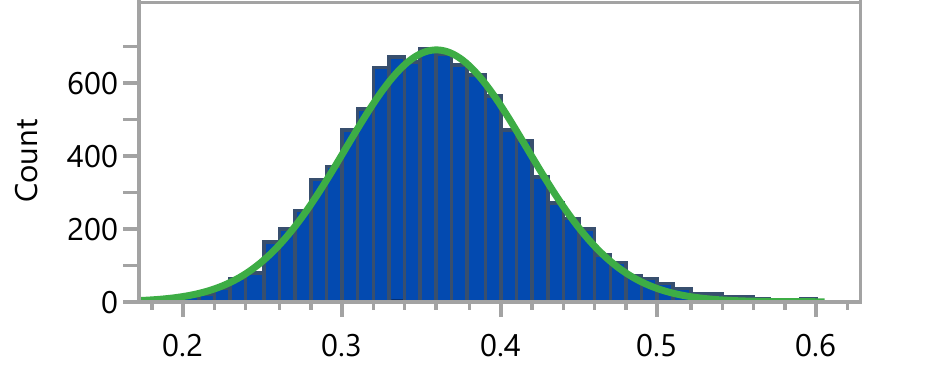}}
		\subfigure[]
		{\label{DistributionAfter}
		\includegraphics[trim = 2cm 21cm 10cm 3.1cm,clip,width=.7\textwidth]{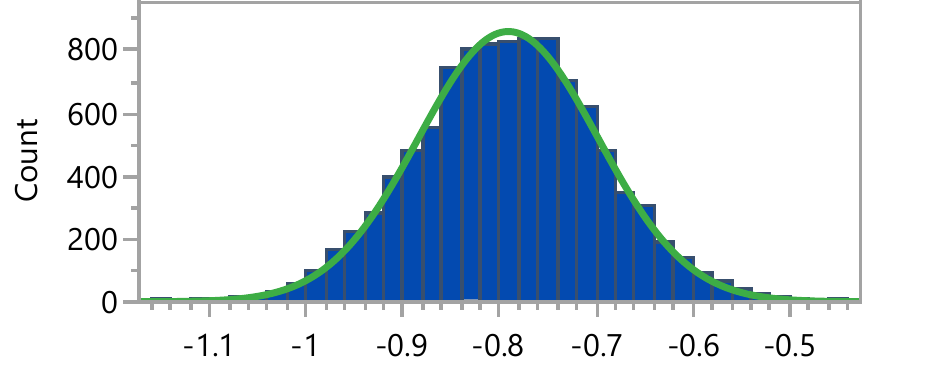}}
	\caption{On the top: a histogram depicting the highest
		incorrect confidence ($\hic{}$) scores of each of 10,000
		perturbed inputs around one of the test points. These scores
		are \emph{not} normally distributed. Beneath: the same
		scores after applying the Box-Cox power transformation, now
		normally distributed.}
\end{figure}

Unfortunately, most often the $\hic{}$ values are not normally distributed. For example, in our
experiments we observed that only 1,282 out of the 10,000 images in the
CIFAR10's test set (fewer than 13\%) demonstrated normally-distributed
$\hic$ values. Figure~\ref{DistributionBefore} illustrates the abnormal
distribution of $\hic$ values of perturbed inputs around one of the
input points. In such cases, we cannot use the normal distribution
function to estimate the probability of adversarial inputs in the
population.

The strategy that we propose for handling abnormal distributions of
data, like the one depicted in Figure \ref{DistributionBefore}, is to
apply \emph{statistical transformations}. Such transformations
preserve key properties of the data, while producing a normally
distributed measurement scale~\cite{GrAmHu13} ---
effectively converting the given distribution into a normal one.
There are two main transformations used to normalize probability distributions:
Box-Cox~\cite{BoCo82} and Yeo-Johnson~\cite{YeJo00}. Here, we focus on
the Box-Cox power transformation, which is preferred for distributions
of positive data values (as in our case). Box-Cox is a continuous,
piecewise-linear power transform function, parameterized by a
real-valued $\lambda$, defined as follows:
\begin{definition}
	\label{definition4}
	The Box-Cox$_\lambda$ power transformation of input $x$ is:
	\[
	\  BoxCox_\lambda (x)=
	\begin{cases}
		\frac{x^{\lambda}-1}{\lambda} & if \lambda \ne 0\\
		\ln (x) & if \lambda=0
	\end{cases}
	\]
\end{definition}
The selection of the $\lambda$ value is crucial for the successful
normalization of the data. There are multiple automated methods for
$\lambda$ selection, which go beyond our scope here~\cite{Ro18}. For our implementation of the technique, we used the
common \emph{SciPy} Python package~\cite{scipy}, which implements one
of these automated methods.

Figure~\ref{DistributionAfter} depicts the distribution of the data
from Figure~\ref{DistributionBefore}, after applying the Box-Cox
transformation, with an automatically calculated $\lambda = 0.534$
value. As the figure shows, the data is now normally distributed:
$\hic(\vec{x}) \sim \mathcal{N} ( \mu=-0.79 , \Sigma=0.092^2)$. The
normal distribution was confirmed with the Anderson-Darling test. Following the Box-Cox transformation,
we can now calculate the Z-Score, which gives $3.71$, and the
corresponding $\plr{}$ score, which turns out to be 99.98\%.

\subsection{The RoMA Certification Algorithm}
\label{section:algorithm}

Based on the previous sections, our method for computing
$\plr{}$ scores is given as Algorithm~\ref{algorithm1}.
The inputs to the algorithm are:
\begin{inparaenum}[(i)]
	\item $\delta$, the confidence threshold for a distinctly adversarial input;
	\item $\epsilon$, the maximum amplitude of perturbation that can be added to $\vec{x_0}$;
	\item $\vec{x_0}$, the input point whose $\plr{}$ score is being computed;
	\item $n$, the number of perturbed samples to generate around $\vec{x_0}$;
	\item $N$, the neural network; and
	\item $\mathcal{D}$, the distribution from which perturbations are drawn.
\end{inparaenum}
The algorithm starts by generating $n$ perturbed inputs around the
provided $\vec{x_0}$, each drawn according to the provided
distribution $\mathcal{D}$ and with a perturbation that does not
exceed $\epsilon$ (lines 1--2); and then storing the $\hic{}$ score of
each of these inputs in the \emph{hic} array (line 3). Next, lines
5--10 confirm that the samples' $\hic{}$ values distribute normally,
applying the Box-Cox transformation if needed. Finally, on lines
11--13, the algorithm calculates the probability of randomly
perturbing the input into a distinctly adversarial input using the
properties of the normal distribution, and returns the computed
$\plr{}_{\delta,\epsilon}(N,\vec{x_0})$ score on line 14.

\begin{figure}[htp]
  \begin{algorithm}[H]
    \caption{Compute Probabilistic Local Robustness($\delta, \epsilon, \vec{x_0}, n, N,
      \mathcal{D}$)}
    \label{algorithm1}
    \begin{algorithmic}[1]
      \FOR {$i:=1$ to $n$} \label{lineRef1} 
      \STATE $\vec{x^i}$ = CreatePerturbedPoint({$\vec{x_0},\epsilon,\mathcal{D}$})
      \STATE $\hic{}$[i] $\gets$ Predict({$N,\vec{x^i}$})
      \ENDFOR \label{lineRef2}
      \IF {Anderson-Darling({hic} $\ne$ NORMAL)} \label{lineRef3} 
      \STATE hic $\gets$ Box-Cox(hic) 
      \IF {Anderson-Darling(hic $\ne$ NORMAL)}
      \STATE Return ``Fail''
      \ENDIF
      \ENDIF\label{lineRef4} 
      \STATE avg $\gets$ Average(hic)\label{lineRef5} 
      \STATE std $\gets$ StdDev(hic)
      \STATE z-score $\gets$ Z-Score(avg,std,$BoxCox(\delta$))\label{lineRef6} 
      \STATE Return NormalDistribution(z-score)\label{lineRef7} 
    \end{algorithmic}
  \end{algorithm}
\end{figure}

\mysubsection{Soundness and Completeness.} Algorithm~\ref{algorithm1}
depends on the distribution of $\hic(\vec{x})$ being normal. If this 
is initially not so, the algorithm attempts to
normalize it using the Box-Cox transformation. The Anderson-Darling
goodness-of-fit test ensures that the algorithm does not treat an
abnormal distribution as a normal one, and thus guarantees the
soundness of the computed $\plr{}$ scores.

The algorithm's completeness depends on its ability to always obtain a
normal distribution. As our evaluation 
demonstrates, the Box-Cox transformation can indeed lead to a normal distribution
very often. However, the transformation might fail in producing a
normal distribution; this failure will be identified by the Anderson-Darling
test, and our algorithm will stop with a failure notice in such
cases. In that sense, Algorithm~\ref{algorithm1} is incomplete.
In practice, failure notices by the algorithm can sometimes be
circumvented --- by increasing the sample size, or by evaluating the
robustness of other input points. 

In our evaluation, we observed that the success of Box-Cox often
depends on the value of $\epsilon$. Small or large $\epsilon$ values
more often led to failures, whereas mid-range values more often led to
success. We speculate that small values of $\epsilon$, which allow
only tiny perturbation to the input, cause the model to assign similar
$\hic{}$ values to all points in the $\epsilon$-ball, resulting in a
small variety of $\hic{}$ values for all sampled points; and
consequently, the distribution of $\hic{}$ values is nearly uniform,
and so cannot be normalized. We further speculate that for large
values of $\epsilon$, where the corresponding $\epsilon$-ball contains
a significant chunk of the input space, the sampling produces a
close-to-uniform distribution of all possible labels, and consequently
a close-to-uniform distribution of $\hic{}$ values, which again 
cannot be normalized. We thus argue that the mid-range values of
$\epsilon$ are the more relevant ones. Adding better support for cases
where Box-Cox fails, for example by using additional statistical
transformations and providing informative output to the user, remains
a work in progress.

\section{Evaluation}
\label{Evaluation}

For evaluation purposes, we implemented Algorithm~\ref{algorithm1} as
a proof-of-concept tool written in
Python 3.7.10, which uses the TensorFlow 2.5 and Keras 2.4 frameworks.
For our DNN, we used a VGG16 network trained for 200 epochs over the
CIFAR10 data set. All experiments mentioned below
were run using the \emph{Google Colab Pro} environment, with an
NVIDIA-SMI 470.74 GPU and a single-core Intel(R) Xeon(R) CPU @
2.20GHz. The code for the tool, the experiments, and the model's
training is available online~\cite{ourCode}.

\begin{figure}[htp]
	\centering
	\includegraphics[trim = 1cm 2.8cm 1cm 2.7cm, clip,width=80mm]{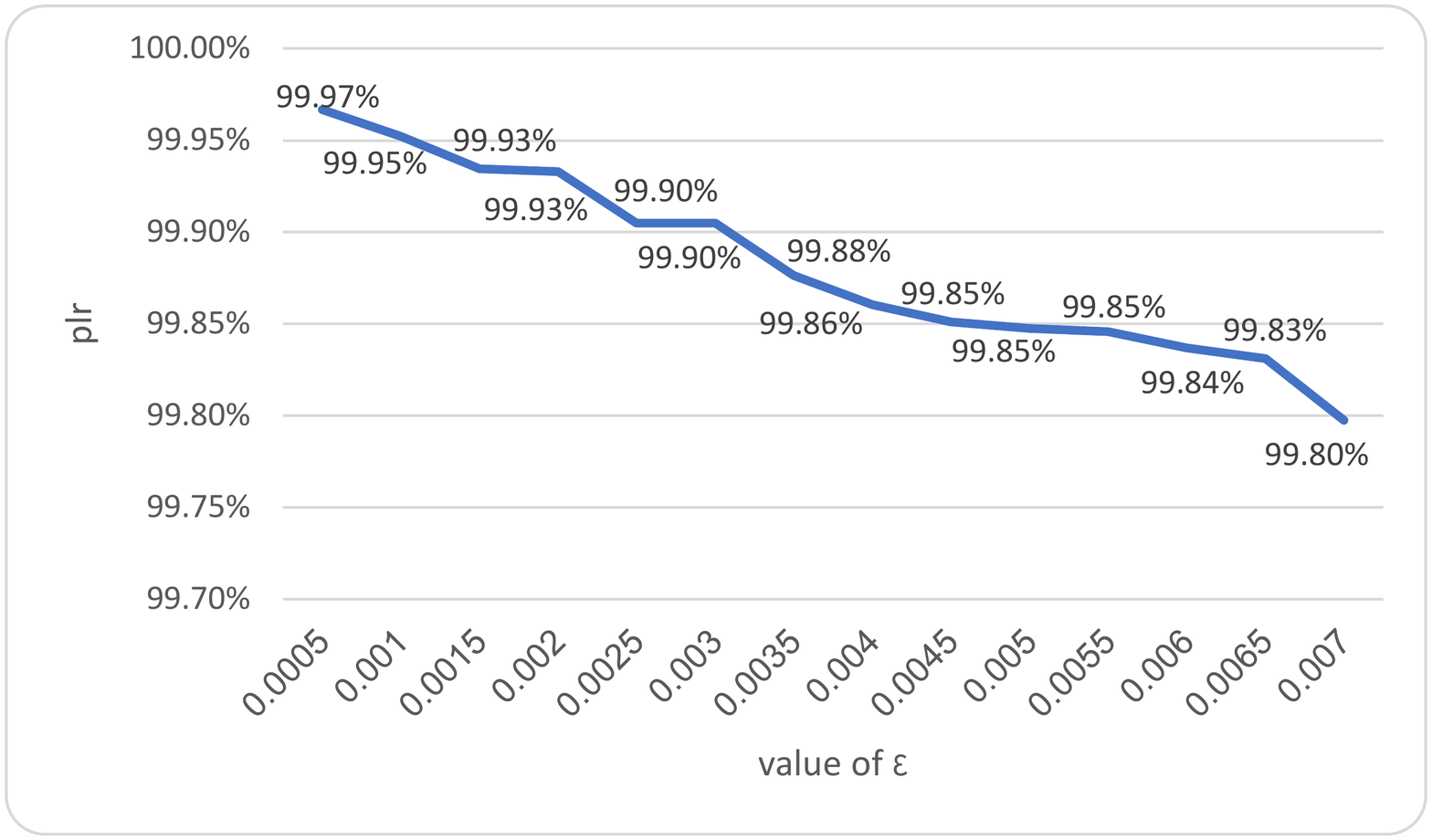} 
	
	\caption{Average $\plr{}$ score of all 10,000 images from the
		CIFAR10 dataset, computed on our VGG16 model as a function
		of $\epsilon$.}
	\label{Figure3}
      \end{figure}

\mysubsection{Experiment 1: Measuring robustness sensitivity to perturbation size.}
By our notion of robustness given in Definition~\ref{definition3}, it
is likely that the $\plr{}_{\delta,\epsilon}(N,\vec{x_0})$ score
decreases as $\epsilon$ increases. For our first experiment, we set
out to measure the rate of this decrease. We repeatedly invoked
Algorithm~\ref{algorithm1} (with $\delta=60\%, n=1,000$) to compute
$\plr{}$ scores for increasing values of $\epsilon$. Instead of
selecting a single $\vec{x_0}$, which may not be indicative, we ran the
algorithm on all 10,000 images in the CIFAR test set, and
computed the average $\plr{}$ score for each value of $\epsilon$; the
results are depicted in Figure~\ref{Figure3}, and indicate a strong 
 correlation between $\epsilon$ and the robustness score. This
result is supported by earlier findings~\cite{WeRaTePa18}. 

Running the experiment took less than 400 minutes, and the algorithm
completed successfully (i.e., did not fail) on 82\% of the queries. We
note here that Algorithm~\ref{algorithm1} naturally lends itself to
parallelization, as each perturbed input can be evaluated
independently of the others; we leave adding these capabilities to our
proof-of-concept implementation for future work.

\begin{figure}[htp]
	\centering
	
	\includegraphics[trim = 1cm 9cm 1cm 10.2cm, clip,width=80mm]{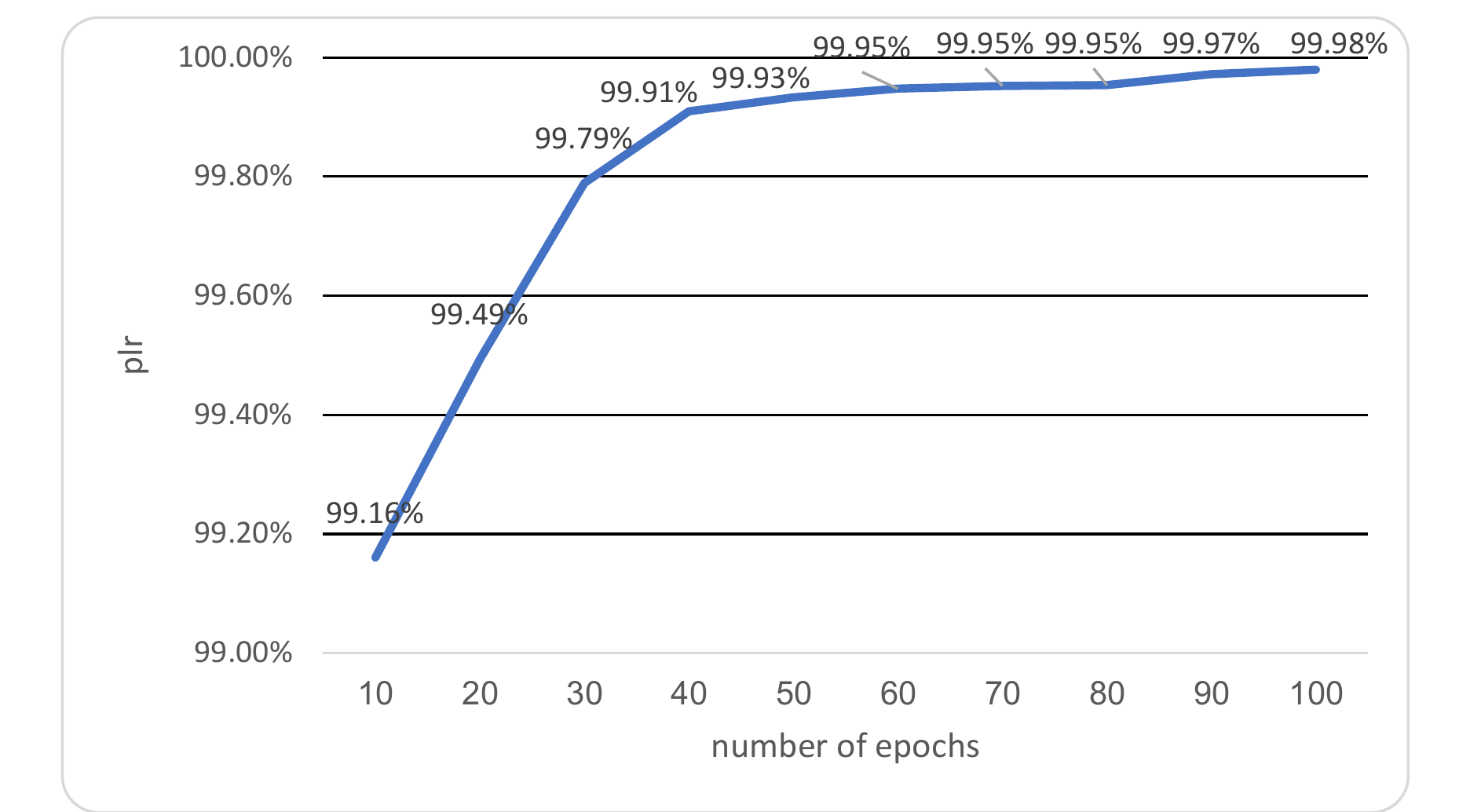} 
	
	\caption{Average \plr{} score of all 10,000 images from
		CIFAR10 test set, computed on VGG16 model as a function of training epochs.}
	\label{Figure5}

      \end{figure}

\mysubsection{Experiment 2: Measuring robustness sensitivity to training epochs.}
In this experiment, we wanted to measure the sensitivity of the
model's robustness 
to the number of epochs in the training process. 
We ran Algorithm~\ref{algorithm1} (with
$\delta=60\%,\epsilon=0.04, n=1,000$) on a VGG16 model trained with a
different number of epochs --- computing the average $\plr{}$ scores
on all 10,000 images from CIFAR10 test set. The computed $\plr{}$
values are plotted as a function of the number of epochs in
Figure~\ref{Figure5}. The results indicate that additional training
leads to improved probabilistic local robustness. These results are also in
line with previous work~\cite{WeRaTePa18}.

\mysubsection{Experiment 3: Categorial robustness.}
For our final experiment, we focused on \emph{categorial robustness}, and
specifically on comparing the robustness scores across categories. We ran Algorithm~\ref{algorithm1} ($\delta$ = 60\%, $\epsilon = 0.04$, and $n=1,000$) on our VGG16 model, for all 10,000 CIFAR10 test set images. The results, divided by category,
appear in Table~\ref{CategorialRobustnes}. For each category we list
the average $plr$ score, the standard deviation of the data (which
indicates the scattering for each category), and the probability of
an adversarial input (the ``Adv'' column,
calculated as $1-plr$). Performing this experiment took 37
minutes. Algorithm~\ref{algorithm1} completed successfully on 90.48\%
of the queries.

The results expose an interesting insight, namely the high variability
in robustness between the different categories. For example, the
probability of encountering an adversarial input for inputs classified
as Cats is four times greater than the probability of encountering an
adversarial input for inputs classified as Trucks. We observe that
the standard deviation for these two categories is very small, which
indicates that they are ``far apart'' --- the difference between Cats
and Trucks, as determined by the network, is generally greater than
the difference between two Cats or between two Trucks. To corroborate this, we applied a
\emph{T-test} and a \emph{binomial test}; and these tests produced a
similarity score of less than 0.1\%, indicating that the two
categories are indeed distinctly different. The important conclusion
that we can draw is that the per-category robustness of models can be
far from uniform.

\newcommand{\fixedmin}[1]{\rule{#1}{0pt}&\rule{#1}{0pt}&\rule{#1}{0pt}&\rule{#1}{0pt}\\[-\arraystretch\normalbaselineskip]}

\begin{figure}[htp]
	\caption{An analysis of average, per-category robustness,
		computed over all 10,000 images from the CIFAR10 dataset.} 
	\label{CategorialRobustnes}
	\begin{center}
		
		\begin{tabular}{llll}\fixedmin{2.5cm}
                            {\bf Category}
			&{\bf  $\plr{}$} 	
			&{\bf  Std-Dev.}
			&{\bf Adv} 
			\\ \hline \\
			
			Airplane	& $99.143\%$ & $5.18\%$ & $0.857\%$ \\
			Automotive 	& $99.372\%$ & $3.86\%$ & $0.628\%$ \\
			Bird 		& $97.226\%$ & $8.87\%$ & $2.774\%$ \\
			Cat			& $97.112\%$ & $8.77\%$ & $2.888\%$ \\
			Deer		& $98.586\%$ & $6.25\%$ & $1.414\%$ \\
			Dog			& $97.233\%$ & $8.58\%$ & $2.767\%$ \\
			Frog		& $98.524\%$ & $6.39\%$ & $1.476\%$ \\
			Horse		& $98.606\%$ & $6.09\%$ & $1.394\%$ \\
			Ship		& $98.389\%$ & $6.63\%$ & $1.611\%$ \\
			Truck		& $99.390\%$ & $4.26\%$ & $0.610\%$ \\
		\end{tabular}
	\end{center}
      \end{figure}
      
It is common in certification methodology to assign each sub-system a
different robustness objective score, depending on the sub-system's
criticality~\cite{FAA93}. Yet, to the best of our knowledge, this is the first time
such differences in neural networks' categorial robustness have been
measured and reported. We believe categorial robustness could affect
DNN certification efforts, by allowing engineers to require separate
robustness thresholds for different categories. For example, for
a traffic sign recognition DNN, a user might require a high robustness
score for the ``stop sign'' category, and be willing to settle for a
lower robustness score for the ``parking sign'' category.

\section{Summary and Discussion}
\label{Discussion}
In this paper, we introduced RoMA --- a novel statistical and scalable
method for measuring the probabilistic local robustness of a black-box, high-scale DNN model. We
demonstrated RoMA's applicability in several aspects. The key advantages of RoMA over
existing methods are:
\begin{inparaenum}[(i)]
	\item it uses a straightforward and intuitive statistical
	method for measuring DNN robustness; 
	\item scalability; and
	\item it works on black-box DNN models, without
	assumptions such as Lipschitz continuity constraints.
\end{inparaenum}
Our approach's limitations stem from the dependence on the normal
distribution of the perturbed inputs, and its failure to produce a
result when the Box-Cox transformation does not normalize the
 data.

The $\plr{}$ scores computed by RoMA indicate the risk of using a DNN
model, and can allow regulatory agencies to conduct \emph{risk
	mitigation} procedures: a common practice for integrating
sub-systems into safety-critical systems. The
ability to perform risk and robustness assessment is an important step
towards using DNN models in the world of safety-critical applications,
such as medical devices, UAVs, automotive, and others. We believe that
our work also showcases the potential key role of \emph{categorial
	robustness} in this endeavor.

Moving forward, we intend to:
\begin{inparaenum}[(i)]
	\item evaluate our tool on additional norms, beyond $L_\infty$; 
	\item better characterize the cases where the Box-Cox transformation
	fails, and search for other statistical tools can succeed in those
	cases; and
	\item improve the scalability of our tool by adding parallelization capabilities.
\end{inparaenum}

\mysubsection{Acknowledgments.} We thank Dr. Pavel Grabov From Tel-Aviv University for his valuable comments and support.

 \bibliographystyle{splncs04}

  \bibliography{mybibliography}

\begin{thebibliography}{10}
\providecommand{\url}[1]{\texttt{#1}}
\providecommand{\urlprefix}{URL }
\providecommand{\doi}[1]{https://doi.org/#1}

\bibitem{AnSo20}
Anderson, B., Sojoudi, S.: {Data-Driven Assessment of Deep Neural Networks with
  Random Input Uncertainty} (2020), {Technical Report.
  \url{http://arxiv.org/abs/2010.01171}}

\bibitem{An11}
Anderson, T.: {Anderson-Darling Tests of Goodness-of-Fit}. Int. Encyclopedia of
  Statistical Science  \textbf{1},  52--54 (2011)

\bibitem{BaIoLaVyNoCr16}
Bastani, O., Ioannou, Y., Lampropoulos, L., Vytiniotis, D., Nori, A.,
  Criminisi, A.: {Measuring Neural Net Robustness with Constraints}. In: Proc.
  30th Conf. on Neural Information Processing Systems (NIPS) (2016)

\bibitem{BeKoSc21}
Berlinger, M., Kolling, S., Schneider, J.: {A Generalized Anderson-Darling Test
  for the Goodness-of-Fit Evaluation of the Fracture Strain Distribution of
  Acrylic Glass}. Glass Structures \& Engineering  \textbf{6}(2),  195--208
  (2021)

\bibitem{BoCo82}
Box, G., Cox, D.: {An Analysis of Transformations Revisited, Rebutted}. Journal
  of the American Statistical Association  \textbf{77}(377),  209--210 (1982)

\bibitem{CaWa17}
Carlini, N., Wagner, D.: {Towards Evaluating the Robustness of Neural
  Networks}. In: Proc. 2017 IEEE Symposium on Security and Privacy (S\&P). pp.
  39--57 (2017)

\bibitem{CoRoZi19}
Cohen, J., Rosenfeld, E., Kolter, Z.: {Certified Adversarial Robustness via
  Randomized Smoothing}. In: Proc. 36th Int. Conf. on Machine Learning (ICML)
  (2019)

\bibitem{DvGaFaKo18}
Dvijotham, K., Garnelo, M., Fawzi, A., Kohli, P.: {Verification of Deep
  Probabilistic Models} (2018), {Technical Report.
  \url{http://arxiv.org/abs/1812.02795}}

\bibitem{EuUnAvSaAg20}
{European Union Aviation Safety Agency}: {Artificial Intelligence Roadmap: A
  Human-Centric Approach To AI In Aviation} (2020),
  \url{https://www.easa.europa.eu/newsroom-and-events/news/easa-artificial-intelligence-roadmap-10-published}

\bibitem{FAA93}
{Federal Aviation Administration}: { RTCA, Inc., Document RTCA/DO-178B }
  (1993), \url{https://nla.gov.au/nla.cat-vn4510326}

\bibitem{GoShSz14}
Goodfellow, I., Shlens, J., Szegedy, C.: {Explaining and Harnessing Adversarial
  Examples} (2014), technical Report. \url{http://arxiv.org/abs/1412.6572}

\bibitem{GrAmHu13}
Griffith, D., Amrhein, C., Huriot, J.M.: {Econometric Advances in Spatial
  Modelling and Methodology: Essays in honour of Jean Paelinck}. Springer
  Science \& Business Media (2013)

\bibitem{Ha13}
Hammersley, J.: {Monte Carlo Methods}. Springer Science \& Business Media
  (2013)

\bibitem{HuHuHuPe21}
Huang, C., Hu, Z., Huang, X., Pei, K.: {Statistical Certification of Acceptable
  Robustness for Neural Networks}. In: Proc. Int. Conf. on Artificial Neural
  Networks (ICANN). pp. 79--90 (2021)

\bibitem{KaBaDiJuKo17}
Katz, G., Barrett, C., Dill, D., Julian, K., Kochenderfer, M.: {Reluplex: An
  Efficient SMT Solver for Verifying Deep Neural Networks}. In: Proc. 29th Int.
  Conf. on Computer Aided Verification (CAV). pp. 97--117 (2017)

\bibitem{KaBaDiJuKo21}
Katz, G., Barrett, C., Dill, D., Julian, K., Kochenderfer, M.: {Reluplex: a
  Calculus for Reasoning about Deep Neural Networks}. Formal Methods in System
  Design (FMSD)  (2021)

\bibitem{LaNi11}
Landi, A., Nicholson, M.: {ARP4754A/ED-79A-Guidelines for Development of Civil
  Aircraft and Systems-Enhancements, Novelties and Key Topics}. SAE Int.
  Journal of Aerospace  \textbf{4},  871--879 (2011)

\bibitem{ourCode}
Levy, N., Katz, G.: {RoMA: Code and Experiments} (2022),
  \url{https://drive.google.com/drive/folders/1hW474gRoNi313G1_bRzaR2cHG5DLCnJl}

\bibitem{MaNoOr19}
Mangal, R., Nori, A., Orso, A.: {Robustness of Neural Networks: A Probabilistic
  and Practical Approach}. In: Proc. 41st IEEE/ACM Int. Conf. on Software
  Engineering: New Ideas and Emerging Results (ICSE-NIER). pp. 93--96 (2019)

\bibitem{MiKwGa2018}
Michelmore, R., Kwiatkowska, M., Gal, Y.: {Evaluating Uncertainty
  Quantification in End-to-End Autonomous Driving Control} (2018), {Technical
  Report. \url{http://arxiv.org/abs/1811.06817}}

\bibitem{PeTh20}
Pereira, A., Thomas, C.: {Challenges of Machine Learning Applied to
  Safety-Critical Cyber-Physical Systems}. Machine Learning and Knowledge
  Extraction  \textbf{2}(4),  579--602 (2020)

\bibitem{Ro18}
Rossi, R.: {Mathematical Statistics: an Introduction to Likelihood Based
  Inference}. John Wiley \& Sons (2018)

\bibitem{scipy}
Scipy: {Scipy Python package} (2021), \url{https://scipy.org}

\bibitem{Ta16}
Taherdoost, H.: {Sampling Methods in Research Methodology; how to Choose a
  Sampling Technique for Research}. Int. Journal of Academic Research in
  Management (IJARM)  (2016)

\bibitem{TiFuRo2021}
Tit, K., Furon, T., Rousset, M.: {Efficient Statistical Assessment of Neural
  Network Corruption Robustness}. In: Proc. 35th Conf. on Neural Information
  Processing Systems (NeurIPS) (2021)

\bibitem{ViGaObOb21}
Vidot, G., Gabreau, C., Ober, I., Ober, I.: {Certification of Embedded Systems
  Based on Machine Learning: A Survey} (2021), {Technical Report.
  \url{http://arxiv.org/abs/2106.07221}}

\bibitem{WeRaTePa18}
Webb, S., Rainforth, T., Teh, Y., Kumar, M.: {A Statistical Approach to
  Assessing Neural Network Robustness} (2018), {Technical Report.
  \url{http://arxiv.org/abs/1811.07209}}

\bibitem{WeChNgSqBoOsDa19}
Weng, L., Chen, P.Y., Nguyen, L., Squillante, M., Boopathy, A., Oseledets, I.,
  Daniel, L.: {PROVEN: Verifying Robustness of Neural Networks with a
  Probabilistic Approach}. In: Proc. 36th Int. Conf. on Machine Learning (ICML)
  (2019)

\bibitem{YeJo00}
Yeo, I.K., Johnson, R.: {A New Family of Power Transformations to Improve
  Normality or Symmetry}. Biometrika  \textbf{87}(4),  954--959 (2000)

\end{thebibliography}

\end{document}